\documentclass{article}
\usepackage{ijcai24}

\usepackage{times}
\usepackage{soul}
\usepackage{url}
\usepackage[hidelinks]{hyperref}
\usepackage[utf8]{inputenc}
\usepackage[small]{caption}
\usepackage{graphicx}
\usepackage{amsmath, amssymb, amsfonts, amsthm}
\usepackage{booktabs}
\usepackage{algorithm}
\usepackage{algorithmic}
\usepackage[switch]{lineno}
\usepackage{color}
\usepackage{subfigure}
\usepackage{multirow}
\usepackage{dsfont}
\theoremstyle{definition}

\author{
Haoyu Chu$^{1,2}$
\and
Yuto Miyatake$^3$\and
Wenjun Cui$^4$\and
Shikui Wei$^1$\And
Daisuke Furihata$^3$\\
\affiliations
$^1$Institute of Information Science, Beijing Jiaotong University\\
$^2$Graduate School of Information Science and Technology, Osaka University\\
$^3$Cybermedia Center, Osaka University\\
$^4$School of Computer and Information Technology, Beijing Jiaotong University
}

\title{Structure-Preserving Physics-Informed Neural Networks \\ With Energy or Lyapunov Structure}

\begin{document}

\maketitle

\begin{abstract}
Recently, there has been growing interest in using physics-informed neural networks (PINNs) to solve differential equations. However, the preservation of structure, such as energy and stability, in a suitable manner has yet to be established. This limitation could be a potential reason why the learning process for PINNs is not always efficient and the numerical results may suggest nonphysical behavior. Besides, there is little research on their applications on downstream tasks. To address these issues, we propose structure-preserving PINNs to improve their performance and broaden their applications for downstream tasks. Firstly, by leveraging prior knowledge about the physical system, a structure‐preserving loss function is designed to assist the PINN in learning the underlying structure. Secondly, a framework that utilizes structure-preserving PINN for robust image recognition is proposed. Here, preserving the Lyapunov structure of the underlying system ensures the stability of the system. Experimental results demonstrate that the proposed method improves the numerical accuracy of PINNs for partial differential equations (PDEs). Furthermore, the robustness of the model against adversarial perturbations in image data is enhanced.
\end{abstract}

\section{Introduction}
\begin{sloppypar}
Physics-informed neural networks (PINNs) \cite{raissi2019physics} aim to embed seamlessly domain knowledge from physical laws into neural networks. Their ability to employ the power of physical principles while leveraging a data-driven approach has shown great potential for addressing complicated tasks in physics, engineering, and beyond \cite{karniadakis2021physics,lu2021deepxde,cuomo2022scientific,liu2023adaptive}. PINNs offer several advantages over conventional numerical methods, such as significantly reducing the computational cost, being mesh-free, and being capable of solving both forward and inverse problems in a unified framework.
\end{sloppypar}

Despite their empirical success, PINNs often suffer from decreased accuracy when dealing with strongly nonlinear and higher-order differential equations \cite{mattey2022novel}. We consider one potential reason is that PINNs overlook the underlying geometric structure of the physical system, leading to numerical solutions that may suggest nonphysical behavior. Most numerical methods have the same issue and therefore fail to preserve the conservative or dissipative property of the dynamical systems. To overcome this limitation, the development of structure-preserving numerical methods has gained momentum \cite{hairer2006structure,sharma2020review,wu2018recent}. A structure-preserving algorithm can guarantee that qualitative characteristics, such as invariant or energy dissipation, are represented in the simulation, providing accurate numerical results over long periods. Thus, it is essential to investigate the application of the structure-preserving idea in PINNs. Nevertheless, most structure-preserving schemes are inapplicable to conventional neural networks because they need an intricate manual derivation of the system equation \cite{matsubara2020deep}. Thanks to PINNs combining differential equations and neural networks, which provides fundamental support for us to introduce the idea of structure-preserving methods into PINNs.

In addition, the current applications of PINNs are mainly focused on solving differential equations. There is little research on their applications on downstream tasks, such as image recognition tasks. Giga \textit{et al.} \shortcite{giga2013behavior} showed the potential power of leveraging the heat equation for binary classification. Another family of  differential equations based deep neural networks, namely neural ordinary differential equations (Neural ODEs) \cite{chen2018neural}, shows promising results on downstream tasks, which describes the continuous dynamics of hidden units utilizing an ODE parameterized by a neural network. The advantage of utilizing PINNs is that one can directly obtain the solutions through the forward inference of the neural network, avoiding the approximation of the numerical integration in Neural ODEs. Besides, we hypothesize that there exists a structure of stability (in the Lyapunov sense) in image recognition tasks, i.e., minor perturbations on the input image will not influence the classification result. We consider the failure of current neural networks on adversarial samples \cite{szegedy2013intriguing,zhang2019adversarial}, crafted by adding minor human-imperceptible perturbations to images, is due to the neglect of preserving the stability structure. Thus, introducing structure-preserving PINNs for image recognition is expected to enable the model to resist initial perturbations. Considering the threat adversarial examples pose to the security of deep learning systems \cite{eykholt2018robust,lu2021towards}, it is meaningful to explore the application of structure-preserving PINNs to robust image recognition. 

To address these problems, we propose a new family of PINNs, named structure-preserving PINNs (SP-PINNs), that can be applied to dynamical systems with energy or Lyapunov structure. Our approach is capable of solving PDEs (e.g., the Allen–Cahn equation) and handling downstream tasks (e.g., image recognition). Our main contributions are summarized as follows: 

\begin{itemize}
  \item Based on prior knowledge about the physical system, we introduce a structure‐preserving loss function to assist PINN in learning the structure of the underlying system, improving its performance on numerical simulation.
  \item We propose an SP-PINN for robust image recognition. Here, we treat the input image after downsampling as the initial value of the differential equation and assume a general form of the unknown underlying dynamical system. In this scenario, preserving the Lyapunov structure of the underlying system ensures the stability of the system. 
  \item Experiments on standard benchmarks show that SP-PINN can consistently outperform the baseline model in terms of robustness against adversarial attacks, which supports our hypothesis on the structure of stability regarding image recognition tasks.
  \item We show that combining the SP-PINN with adversarial training methods further enhances the robustness against adversarial examples.
\end{itemize}

\section{Related Works}

\subsection{Neural Networks for Differential Equations}

The first attempts at utilizing neural networks to solve differential equations began in the 1990s \cite{lee1990neural,meade1994solution,lagaris1998artificial}. Recent progress in automatic differentiation techniques has enabled researchers to design more complicated neural network architectures.  

Among different data-driven techniques for solving differential equations, PINNs \cite{raissi2019physics} have shown remarkable promise and versatility. Mattey \textit{et al.} \shortcite{mattey2022novel} designed a novel PINN scheme that re-trains the same neural network for solving the PDE over successive time segments while satisfying the already obtained solution for all previous time segments. Krishnapriyan \textit{et al.} \shortcite{krishnapriyan2021characterizing} analyzed that possible failure modes in PINNs are attributed to the PINNs’ setup making the loss landscape very hard to optimize.  \cite{jagtap2021extended} extended PINN by leveraging the generalized space-time domain decomposition for solving arbitrary complex-geometry domains. Wang \textit{et al.} \shortcite{wang2024pinn} proposed a neural architecture search-guided PINN for solving PDEs.

\subsection{Structure-Preserving Deep Learning}

Greydanus \textit{et al.} \shortcite{greydanus2019hamiltonian} parameterized the Hamiltonian mechanics with a neural network and then learned it via a data-driven approach, which conserved an energy-like quantity. Sosanya \textit{et al.} \shortcite{sosanya2022dissipative} proposed a dissipative Hamiltonian neural network that leverages the tools of Hamiltonian mechanics and Helmholtz decomposition to separate conserved quantities from dissipative quantities. Lutter \textit{et al.} \shortcite{LutRitPet19} proposed a neural network for learning the mechanic systems of the Euler-Lagrange equation employing end-to-end training while keeping physical plausibility. \cite{matsubara2023finde} utilized Neural ODEs for finding and preserving invariant quantities by leveraging the projection method and the discrete gradient method \cite{matsubara2020deep}. Jagtap \textit{et al.} \shortcite{jagtap2020conservative} presented a conservative PINN on discrete sub-domains by using a separate PINN in each sub-domain, and then stitching back all sub-domains through the corresponding conservative quantity, which is different from our approach that is based on the prior knowledge about the underlying dynamics and is theoretically applicable both conservative and dissipative systems.

\subsection{Adversarial Defense Methods}

Since realizing the instability of deep neural networks, researchers have proposed different kinds of complementary techniques to defend adversarial examples, such as distillation defense \cite{papernot2016distillation,chu2023improving} and adversarial training. Maday \textit{et al.} \shortcite{madry2017towards} proposed an adversarial training method by injecting adversarial examples generated by PGD attacks into training data. Ilyas \textit{et al.} \shortcite{ilyas2019adversarial} created a robust dataset for adversarial training by removing non-robust features from the dataset. Lamb \textit{et al.} \shortcite{LAMB2022218} augmented the adversarial training with interpolation-based training, which aims to tackle the problem that traditional adversarial training aggravates the generalization performance of the networks on clean data.

After knowing the connection between dynamical systems and deep neural networks, designing defense methods based on the Lyapunov stability theory becomes a new trend. Kang \textit{et al.} \shortcite{kang2021stable} proposed stable Neural ODE based on Lyapunov’s first method for resisting adversarial examples. Chu \textit{et al.} \shortcite{chu2023learning} leveraged Lyapunov’s second method for preventing successful adversarial attacks by inherently constraining the deep equilibrium model \cite{bai2019deep} to be stable.

It is worth mentioning that our proposed method is the first attempt at utilizing modified PINNs for adversarial defense.

\section{Methodology}

This section presents the applications of the proposed SP-PINNs to PDEs (e.g., the Allen–Cahn equation), and robust image recognition. 

The strategies are summarized as follows: firstly, a neural network is specified for each of the different tasks to fit the mapping from the input to the numerical solution, and then the derivatives w.r.t. the input is calculated by automatic differentiation; secondly, the core of the proposed method involves incorporating prior knowledge about the energy or Lyapunov structure of the system into the specific neural network; thirdly, the neural network is trained by minimizing the loss function. For different scenarios, we design a corresponding training process and loss function to cater to the unique characteristics of each problem domain.

\subsection{SP-PINN for PDEs}
\label{sp_pde}

We consider the Allen–Cahn equation \cite{allen1972ground}, which is a strongly nonlinear reaction-diffusion equation. The Allen–Cahn equation is widely used for modeling some phase separation and domain coarsening phenomena. Currently, PINN's accuracy suffers significantly from strongly nonlinear PDEs \cite{mattey2022novel}. Therefore, we aim to improve the solutions' accuracy by learning the underlying structure of the equation. The Allen–Cahn equation is formulated as:
\begin{equation}
    \frac{\partial{u}}{\partial{t}} = pu + ru^3 + q\frac{\partial{^2 u}}{\partial{x}^2},
\end{equation}
where $p > 0, q > 0, r < 0$, $x \in [0, L]$ and $t\in [0, T]$. The initial condition is $u(x,0) = u_0(x)$. Here, we employ the Neumann boundary condition $u_x(0,t)=u_x(L,t)$.

To reveal the energy dissipative property of the Allen–Cahn equation, we first introduce a quality, which is called the `free energy' or `local energy' of the problem:
\begin{equation}
G(u,u_x) = -\frac{p}{2}u^2 -\frac{r}{4}u^4 + \frac{q}{2}u_{x}^2.
\end{equation}

The evolution of the solution is shown to evolve in a direction that the global energy is decreased \cite{furihata2010discrete}:
\begin{equation}
\frac{d}{dt}J(u) = \frac{d}{dt} \int_{0}^{L} G(u,u_x)dx \leq 0,
\end{equation}
where $J(u)$ is the `global energy', which is a functional of $u$. Meanwhile, $J(u)$ can be regarded as a function of $t$.

\begin{figure}[t]
\begin{center}
\includegraphics[width=0.47\textwidth]{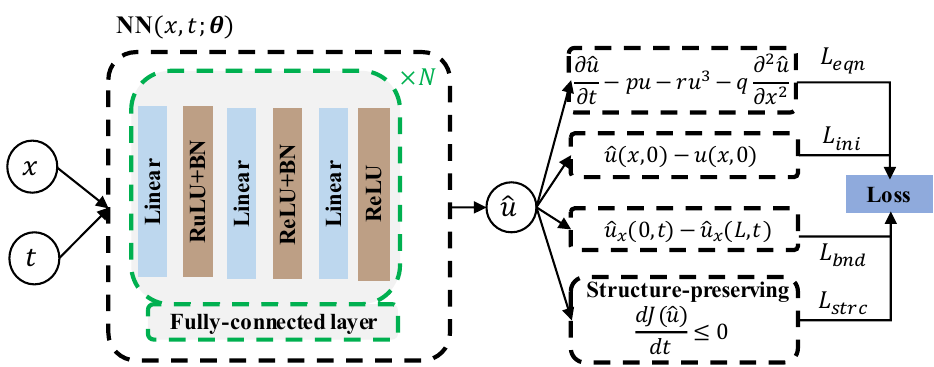}
\end{center}
\caption{SP-PINN for Allen–Cahn equation. The SP-PINN is defined as a $N$-layered FCN with inputs $x,t$ and output $\hat{u}(x,t)$.}
\label{fig:pinn_ac}
\end{figure}

\begin{sloppypar}
To solve the Allen–Cahn equation, the structure-preserving PINN is defined as a $N$-layer fully connected network (FCN) with inputs $x,t$ and output $\hat{u}(x,t)$. The framework is illustrated in Figure \ref{fig:pinn_ac}. 

We obtain the derivatives of the output $\hat{u}(x,t)$ w.r.t. position $x$ and time $t$ via automatic differentiation and randomly select $N_f$ in the spatiotemporal region $(x,t)$ to calculate the equation residual:
\begin{equation}
L_{\text{eqn}} = \frac{1}{N_f}\sum^{N_f}_{i=1} \|  \frac{\partial{\hat{u}_i}}{\partial{t}} - p\hat{u}_i - r \hat{u}^{3}_i - q\frac{\partial{^2 \hat{u}_i}}{\partial{x}^2}\|_2 .
\end{equation}
\end{sloppypar}

Then, we randomly select $N_i$ points in the region $(x,0)$ to calculate the initial condition residual:
\begin{equation}
L_{\text{ini}} = \frac{1}{N_i}\sum^{N_i}_{i=1} \| \hat{u}_{i} - u_i \|_2.
\end{equation}

Next, we randomly select $N_b$ points in the regions $(0,t)$ and $(L,t)$ to calculate the boundary condition residual:
\begin{equation}
L_{\text{bnd}} = \frac{1}{N_b}\sum^{N_b}_{i=1} \| \hat{u}^i_{x}(0) - \hat{u}^i_{x}(L)  \|_2 .
\end{equation}

Finally, to specify the structural loss, we define the discrete global energy accordingly by:
\begin{equation}
J(\hat{u}) \triangleq {\sum_{k=0}^M}  {^{\prime \prime}G}_{k}(\hat{u}, \hat{u}_x)\Delta x,
\end{equation}
where $\Delta x = L/M$, $M$ is the number of the spatial grid points, ${\sum_{k=0}^M}  {^{\prime \prime}f}$ denotes the trapezoidal rule:
\begin{equation}
{\sum_{k=0}^M}  {^{\prime \prime}f}  \triangleq \frac{1}{2}f_0 + f_1+ \cdots + f_{M-1} + \frac{1}{2}f_M.
\end{equation}

We uniformly select $N_e$ points in the time interval $[0, T]$ to calculate the structural loss:
\begin{equation}
L_{\text{strc}}=\frac{1}{N_e}\sum^{N_e}_{i=1}\|\operatorname{ReLU}(\frac{d}{dt} J(\hat{u}_{i}))\|_2,
\end{equation}
where $\operatorname{ReLU} = \operatorname{max}(0,x)$ is the rectified linear unit function.

Therefore, the total loss function for training the proposed model is the summation of the equation residual, the initial condition residual, the boundary condition residual, and the structural loss:
\begin{equation}
L_{\text{total}}  = \lambda_1 L_{\text{eqn}} +\lambda_2 L_{\text{bnd}} + \lambda_3 L_{\text{ini}} + \lambda_4 L_{\text{strc}},
\end{equation}
where $\lambda_1, \lambda_2, \lambda_3, \lambda_4$ are hyperparameters.

\begin{figure*}[t]
\begin{center}
\includegraphics[width=0.9\textwidth]{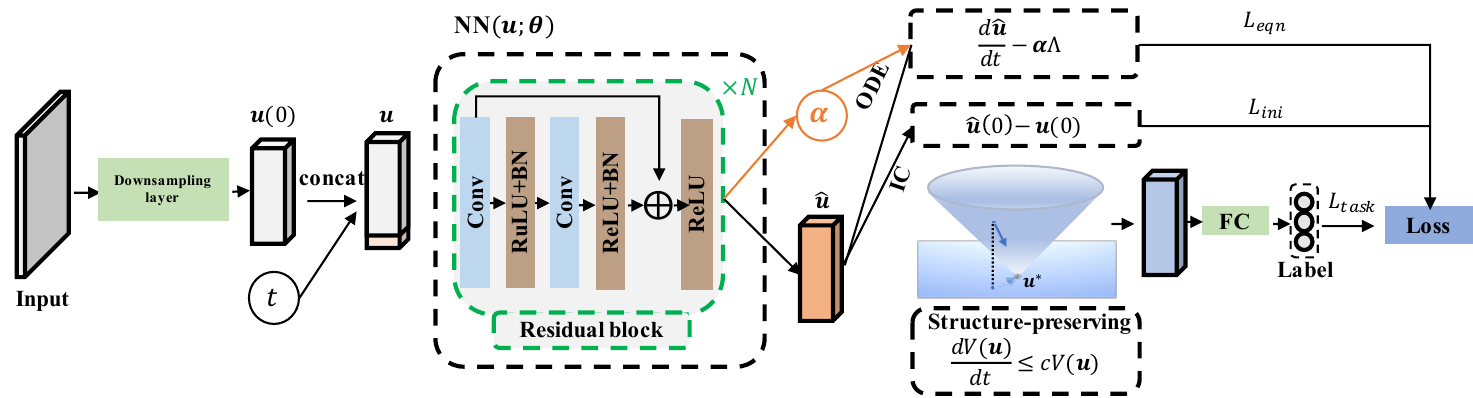}
\end{center}
\caption{SP-PINN for robust image recognition. The SP-PINN is defined as a $N$-layer residual network. The orange line denotes the process of solving the inverse problem, in which a data-driven way is resorted to obtain the unknown parameter $\boldsymbol{\alpha}$ of the underlying system. For solving the forward problem, the output is the approximate solution $\boldsymbol{\hat{u}}$. The blue arrow represents a state that satisfies the Lyapunov exponential stability condition. The FC layer predicts the category of the image. When calculating the loss regarding the initial condition, the time $t$ is set to 0. When performing classification, the time $t$ is set to 1.}
\label{fig:pinn_images}
\end{figure*}

\subsection{SP-PINN for Robust Image Recognition}

Currently, PINNs are mostly utilized for solving differential equations. In this section, we delve into the extension of PINNs to downstream tasks and propose an SP-PINN for robust image recognition. First, we define the image recognition task as an initial value problem, where the input image after downsampling is treated as the initial value, and the approximate solution is obtained through the evolution of time. Secondly, we project the learned dynamical system into a space that satisfies the Lyapunov stability condition to preserve the stability structure.

\subsubsection{Problem Statement}

We hypothesize that there is a structure of stability in the Lyapunov sense in image recognition tasks, that is, small changes put on the input image will not influence the classification result. Thus, introducing the structure-preserving idea into the image recognition tasks is expected to enable the classification model to have resistance against the minor perturbations added on the input image, making the model more robust. 

However, the challenge of utilizing PINN for image recognition is that the underlying dynamical system is unknown. Therefore, we specify the following ODE to describe the system:

\begin{equation}
\frac{d\boldsymbol{u}(t)}{dt} = f(\boldsymbol{u}(t), t; \boldsymbol{\theta}),
\label{eqn:ode_system}
\end{equation}
where the initial value is $\boldsymbol{u}(0)=\boldsymbol{u}_0$ and the time interval is $[0, T]$.

It is worth mentioning that Neural ODEs \cite{chen2018neural} also use Eqn.~(\ref{eqn:ode_system}) to describe the continuous dynamics of hidden units, which show promising results on downstream tasks. Our proposed method diverges from Neural ODEs in terms of the approach to solving the initial value problem. 

Theoretically, the analytic solution can be obtained by integrating Eqn. (\ref{eqn:ode_system}):
\begin{equation}
\boldsymbol{u}(T) = \boldsymbol{u}(0) + \int_{0}^{T} f(\boldsymbol{u}(t), t) \,dt.
\end{equation}
In practice, Neural ODEs utilize the ODE solver, such as Runge–Kutta, Dopri5, to approximate the numerical integration $\int_{0}^{T} f(\boldsymbol{u}(t), t) \,dt$. This procedure is formulated by:
\begin{equation}
\boldsymbol{u}(T) = \operatorname{ODESolver}((\boldsymbol{u}(0), f, [0, T], \boldsymbol{\theta})).
\end{equation}

Unlike Neural ODEs, the advantage of utilizing PINN is that PINN can obtain the numerical solution at time $T$ directly through the forward inference of the neural network, that is, $\boldsymbol{u}(T) = \operatorname{PINN}((\boldsymbol{u}(0), T, \boldsymbol{\theta}))$, avoiding the approximation of the numerical integration. Nevertheless, this also poses a problem in that the form of Eqn. (\ref{eqn:ode_system}) needs to be known in advance. To address this problem, we assume a general form of the underlying dynamical system following Hu \textit{et al.} \shortcite{hu2022revealing}:
\begin{equation}
\begin{aligned}
\frac{d\boldsymbol{u}(t)}{dt}& = \boldsymbol{\alpha} \boldsymbol{\Lambda}, \\
\boldsymbol{u}(0)&= (u_1(0), u_2(0), \cdots , u_n(0))^T,
\label{eqn:ode_image}
\end{aligned}
\end{equation}
where $\boldsymbol{\Lambda} = \{1, u_1, u_2, \cdots, u_n, u_{1}u_{1}, u_{1}u_{2}, \cdots, u^{p}_{n}\}$ is the complete set of $p^{th}$-order polynomial basis with unknown coefficients $\boldsymbol{\alpha}=(\alpha_{ij})_{n \times M}$, $M=\binom{p+n}{p}$. Here, we take the downsampling result of the input image as the initial value $\boldsymbol{u}(0)$. Like Neural ODEs, the final time for PINN is set to $T=1$. 

To obtain the unknown parameters $\boldsymbol{\alpha}$, we need to first solve the inverse problem. As long as we know the underlying dynamical system, we can calculate the numerical solution via the forward inference, i.e., solving the forward problem. Motivated by Raissi \textit{et al.} \shortcite{raissi2019physics} and Kim \textit{et al.} \shortcite{kim2023partial}, the learning process of the proposed model is achieved by solving the inverse and forward problems alternately.

\subsubsection{Details on the Learning Process}

The SP-PINN for image recognition is defined as a $N$-layer residual network, where the input $\boldsymbol{u}$ is a vector concatenated by initial value $\boldsymbol{u}(0)$ with time $t$ (after broadcasting). The scratch of the architecture of the proposed method is shown in Figure \ref{fig:pinn_images}. 

\textbf{How to solve the inverse problem?} We solve the inverse problem through a data-driven approach. After initializing the network parameters, we can input the images of the training set and get their corresponding outputs. Then, we minimize the following loss function regarding equation residual to find the $\boldsymbol{\alpha}$:
\begin{equation}
\label{eqn:inverse}
\mathop{\arg\min}_{ \boldsymbol{\alpha}} \frac{1}{N_{t}}\sum^{N_{t}}_{i=1} \| \frac{d \boldsymbol{\hat{u}}_{i}}{dt} - \boldsymbol{\alpha} \boldsymbol{\Lambda}\|,
\end{equation}
where $N_{t}$ is the number of the training set.

\textbf{How to solve the forward problem?} As long as the underlying system is known, we can calculate the approximate solutions through the forward inference. The new solutions are in turn utilized for solving the inverse problem to update $\boldsymbol{\alpha}$. Repeated iterations of this process yield more accurate $\boldsymbol{\alpha}$ and $\boldsymbol{\hat{u}}$.

While solving the inverse problem, we fix the trainable parameters $\boldsymbol{\theta}$ of the neural network and learn the unknown parameters $\boldsymbol{\alpha}$ of the underlying dynamical system. While solving the forward problem, we fix the parameters $\boldsymbol{\alpha}$ of the dynamical system and train the parameters $\boldsymbol{\theta}$ of the neural network, in this case, the output of the model is the approximate solution $\boldsymbol{\hat{u}}$. 

\textbf{How to preserve the stability structure?} To preserve the stability structure, that is, ensuring the proposed model is robust to minor perturbations on the initial value, we jointly learn a convex positive definite Lyapunov function along with dynamics constrained to be stable and project the learned dynamical system of PINN onto a space where the Lyapunov exponential stability condition \cite{giesl2015review} holds. Please refer to \cite{giesl2015review} for more details on the definition of Lyapunov stability and the Lyapunov stability theorem.  

We construct the structure-preserving module motivated by Manek and Kolter \shortcite{kolter2019learning}. Let $F(\boldsymbol{u}) = d \operatorname{NN}(\boldsymbol{u})/ dt$ be a basic dynamic model, let $V: \mathds{R}^n \rightarrow \mathds{R}$ be a positive definite function, and $c$ be a nonnegative constant, the exponentially stable dynamical model is defined as
\begin{equation}
    \begin{aligned}
    \widetilde{F}(\boldsymbol{u}) & =\operatorname{Projection}(F(\boldsymbol{u}),\{F: \nabla V(\boldsymbol{u})^T F \leq- c V(\boldsymbol{u})\}) \\
    &= \begin{cases}F(\boldsymbol{u}) & \text { if } \phi(\boldsymbol{u}) \leq 0 \\ 
    F(\boldsymbol{u})-\nabla V(\boldsymbol{u}) \frac{\phi(\boldsymbol{u})}{\|\nabla V(\boldsymbol{u})\|_2^2} & \text { otherwise }\end{cases},
    \end{aligned}
\end{equation}
where, $\phi(\boldsymbol{u})= \nabla V(\boldsymbol{u})^T F(\boldsymbol{u}) + c V(\boldsymbol{u})$. For $\phi(\boldsymbol{u}) > 0$, the output of the base dynamics model is projected onto a halfspace where this condition holds, otherwise, the output is returned unchanged. 

The Lyapunov function $V$ is defined as positive definite and continuously differentiable, and has no local minima:
\begin{equation}
V(\boldsymbol{u})=\sigma(g(\boldsymbol{u})-g(0))+\eta\|\boldsymbol{u}\|_2^2,
\end{equation}
where $\sigma$ is a positive convex non-decreasing function with $\sigma_{k}(0) = 0$,  $\eta$ is a small constant, and $g$ is represent as a input-convex neural network (ICNN)~\cite{amos2017input}. Since the Lyapunov function is defined as continuously differentiable, we can use automatic differentiation to compute its gradient. This advantageous feature allows us to train our final network in a manner similar to any other network.

Consequently, we ensure that the proposed model satisfies the Lyapunov exponentially stable condition through the procedures described above, thereby preserving the Lyapunov structure of the underlying dynamical system. 

The FC layer, that is, a weighted linear transformation, takes part in predicting the category of the image.

The loss function for training the SP-PINN is obtained as follows. First, obtain the derivative of the network's output $\boldsymbol{\hat{u}}$ w.r.t. time $t$. Second, calculate the equation residual:
\begin{equation}
L_{\text{eqn}} = \frac{1}{N_{t}}\sum^{N_{t}}_{i=1} \| \frac{d \boldsymbol{\hat{u}}_{i}}{dt} - \boldsymbol{\alpha} \boldsymbol{\Lambda}\| .
\label{eqn:image_eqn}
\end{equation}
As can be seen, the equation residual in the forward problem is the same form as in the inverse problem, i.e., Eqn. (\ref{eqn:inverse}). Hence, PINNs can deal with forward and inverse problems in a unified paradigm, which offers a great advantage compared to traditional numerical methods \cite{chirigati2021inverse} that need to design different schemes for different inverse problems.

Then, we calculate the initial condition residual:
\begin{equation}
L_{\text{ini}} = \frac{1}{N_h}\sum^{N_h}_{i=1} \| \boldsymbol{\hat{u}}_{i}(0) - \boldsymbol{u}_{i}(0)  \|_2,
\label{eqn:image_ini}
\end{equation}
where ${N_h}$ is the number of elements in $\boldsymbol{u}_{i}(0)$.

Finally, we utilize the cross-entropy (CE) loss to measure the difference between the FC layer's result $\boldsymbol{y}$ and the true label $\boldsymbol{\hat{y}}$ of the image:
\begin{equation}
L_{\text{task}} = \frac{1}{N_{t}} \sum^{N_{t}}_{i=1} \operatorname{CE}(\boldsymbol{y}_{i}, \boldsymbol{\hat{y}}_{i}), 
\end{equation}

The loss function used to update the network parameters $\boldsymbol{\theta}$ is the summation of the equation residual, the initial condition residual, and the cross-entropy loss. Therefore, the best parameters $\boldsymbol{\theta}$ is obtained by minimizing the loss function:
\begin{equation}
\mathop{\arg\min}_{ \boldsymbol{\theta}} \ \lambda_1 L_{\text{eqn}} + \lambda_2 L_{\text{ini}} + \lambda_3 L_{\text{task}},
\end{equation}
where $\lambda_1, \lambda_2, \lambda_3$ are hyperparameters.

\section{Experiments}

In this section, we first present the experimental setup. Then, we evaluate the proposed method on two different scenarios and analyze the experimental results. 

\begin{figure*}[t]
\begin{center}
\includegraphics[width=0.9\textwidth]{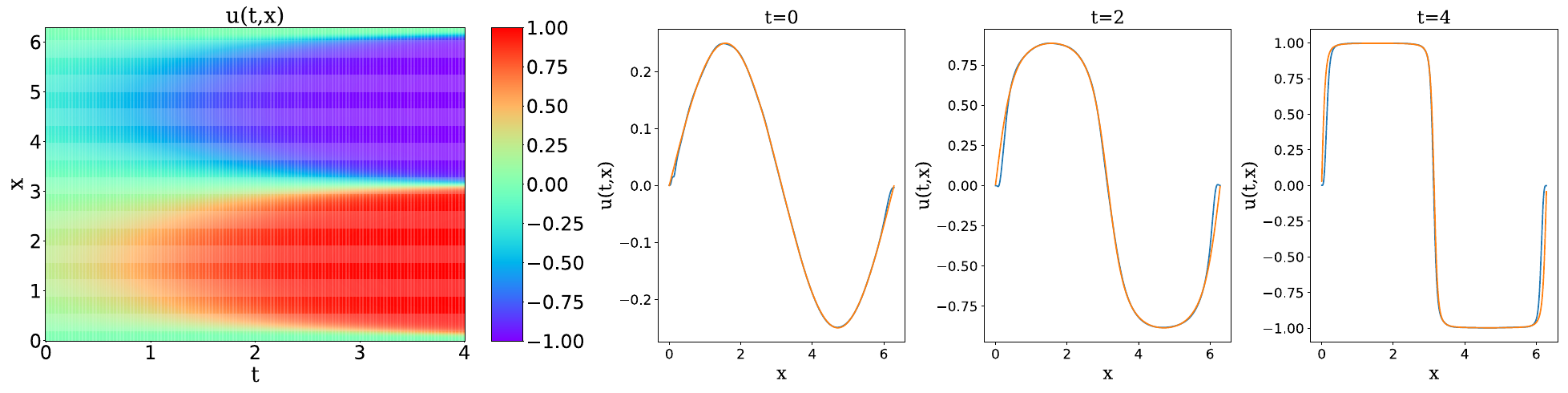}
\end{center}
\caption{Numerical solutions of the Allen–Cahn equation. The orange line is obtained by DVDM. The blue line is obtained by our method.}
\label{exp:pde1}
\end{figure*}

\begin{figure}[t]
\begin{center}
\includegraphics[width=0.49\textwidth]{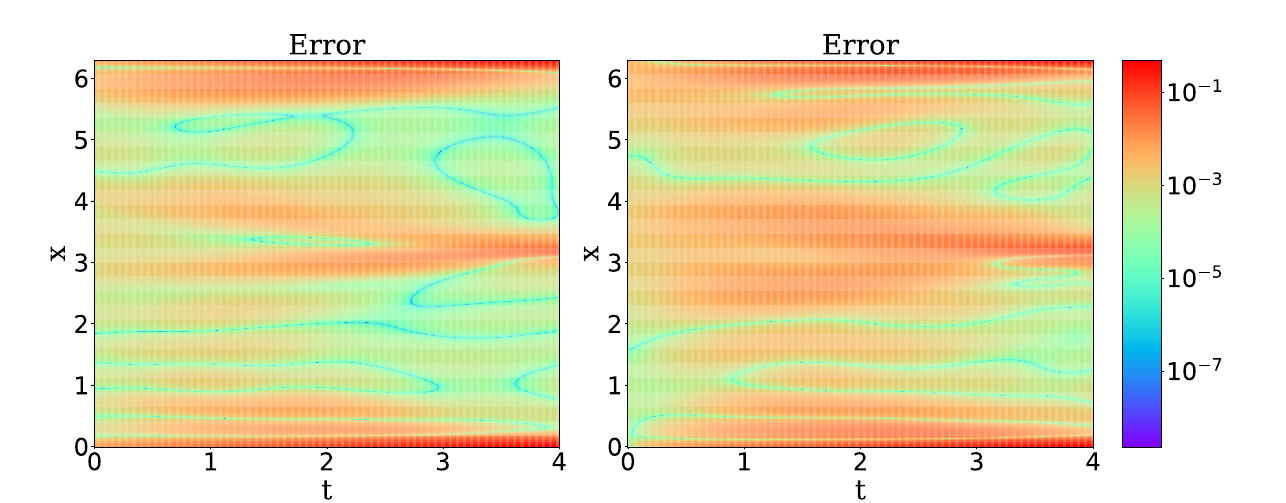}
\end{center}
\caption{(left) The error between the results obtained by the proposed model and DVDM. (right) The error between the results obtained by the baseline model and DVDM. }
\label{exp:pde2}
\end{figure}

\subsection{Experimental Setup}

We use PyTorch~\cite{paszke2017automatic} framework for the implementation. The torch version is 1.11.0+cu113. We conducted our experiments on the Ubuntu 20.04.6 LTS operator system. All the experiments are run on a single NVIDIA A100 40GB GPU. In our experiments, we set all the hyperparameters $\lambda_i$ as 1. 

\subsection{Experiments on the Allen–Cahn Equation}

Regarding the training configurations, we first run the Adam algorithm \cite{kingma2014adam} with 10,000 epochs and then employ the limited-memory BFGS algorithm \cite{liu1989limited}.

We show a numerical example in Figure \ref{exp:pde1} with $p = 1, r=-1, q=0.0001$ and $x \in [0, 2\pi]$, $t\in [0,4]$. The initial condition is taken to $u_0(x) = 0.25 sin(x)$ and we employ the Neumann boundary condition $u_x(0,t)=u_x(2\pi,t)$. In this experiment, the $N_f$, $N_b$, $N_i$, $N_e$ is set to 8,000, 1,000, 1,000, 2000. $\Delta x$ is $L/M = 2\pi / 2000 \approx 0.00314$. We employ an FCN with 6 hidden layers. 

\begin{figure}[t]
\begin{center}
\includegraphics[width=0.4\textwidth]{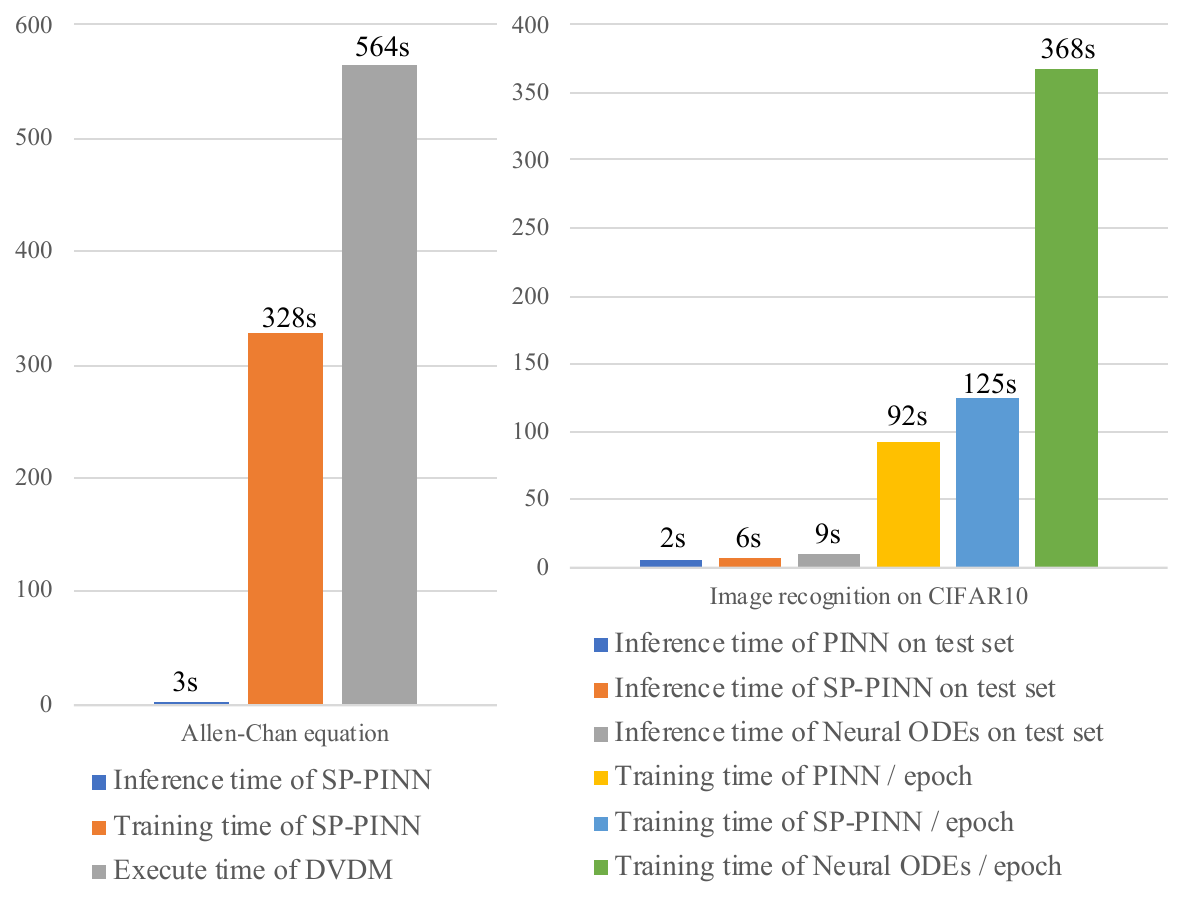}
\end{center}
\vspace{-1em}
\caption{The comparison of the time consuming on same experimental settings. The ODE solver used in Neural ODEs is Dopri5.}
\label{exp:time}
\end{figure}

We compare the proposed method with the discrete variational derivative method (DVDM) \cite{furihata2010discrete}. DVDM is a structure-preserving numerical method for PDEs, which improves the qualitative behavior of the PDE solutions and allows for stable computing. The procedure of DVDM is detailed in \cite{furihata2010discrete}. Although SP-PINN is slightly inferior to DVDM in terms of accuracy, DVDM needs an intricate derivation to construct a specific numerical scheme. Besides, Figure \ref{exp:time} illustrates that SP-PINN obtains numerical solutions significantly faster than DVDM. Even accounting for the training time of the network, SP-PINN is less time-consuming than DVDM, demonstrating the effectiveness of the SP-PINN.

As shown in Figure \ref{exp:pde2}, the error between the results obtained by the proposed model and DVDM is smaller than that of the baseline model (i.e., the vanilla PINN) and DVDM, which indicates that the numerical solutions of the proposed model are more accurate than that of the baseline model.

\subsection{Experiments on Image Recognition}

\subsubsection{The Experimental Configurations}

We conduct a set of experiments on four datasets, MNIST \cite{lecun1998gradient}, Street View House Numbers (SVHN) \cite{yuval2011reading} and CIFAR10/100 \cite{krizhevsky2009learning}. 


We choose a 2-layer ICNN and set the activation function $ \sigma$ as a smooth ReLU function. $\eta$ is set to 0.001. For optimization, we use the Adam algorithm with the initial learning rate $= 0.001$ and a cosine annealing schedule. The training epochs for MNIST, SVHN, and CIFAR10/100 are set to 10, 40, and 60/70.

For MNIST, we downsample the input image from 28$\times$28 to 6$\times$6. For SVHN and CIFAR, we downsample the input image from 32$\times$32 to 16$\times$16. We choose an 18-layer residual network \cite{he2016deep} as the backbone of the SP-PINN and utilize a Maclaurin series with $5^{th}$-order polynomial basis for modeling the underlying systems.

We test the performance of the PINN and SP-PINN on two white-box adversarial attacks: iterative fast gradient sign method (I-FGSM) ~\cite{kurakin2018adversarial} and project gradient descent (PGD) \cite{madry2017towards}. The configurations for I-FGSM and PGD is detailed in the supplementary material.

\begin{table*}[t]
\begin{tabular}{llllllll}
\hline
Benchmark                 & Model                            & Clean                  & Attack & $\epsilon=2/255$       & $\epsilon=4/255$        & $\epsilon=6/255$        & $\epsilon=8/255$        \\ \hline
\multirow{4}{*}{MNIST}    & \multirow{2}{*}{PINN (baseline)} & \multirow{2}{*}{99.40}      & I-FGSM & 94.72                  & 92.91                   & 90.55                   & 87.82                   \\
                          &                                  &                        & PGD    & 94.88                  & 92.99                   & 90.81                   & 87.89                   \\
                          & \multirow{2}{*}{SP-PINN }  & \multirow{2}{*}{99.38} & I-FGSM & \textbf{98.77} (+4.05) & \textbf{98.75} (+5.84)  & \textbf{98.67} (+8.12)  & \textbf{98.49} (+10.67) \\
                          &                                  &                        & PGD    & \textbf{98.76} (+3.88) & \textbf{98.75} (+5.76)  & \textbf{98.72} (+7.91)  & \textbf{98.56} (+10.67) \\ \hline
\multirow{4}{*}{SVHN}     & \multirow{2}{*}{PINN (baseline)} & \multirow{2}{*}{93.80}      & I-FGSM &       64.47                 &         59.43                &               56.39          &          53.47               \\
                          &                                  &                        & PGD    &    61.67                    &     56.80                    &     54.00                    &    50.71                     \\
                          & \multirow{2}{*}{SP-PINN}  & \multirow{2}{*}{92.69}      & I-FGSM &     \textbf{66.54} (+2.07)                  &     \textbf{65.44} (+6.01)                  &         \textbf{64.26} (+7.87)              &    \textbf{62.74} (+9.27)                    \\
                          &                                  &                        & PGD    &    \textbf{65.54} (+3.87)                   &    \textbf{63.61} (+6.81)                    &         \textbf{61.97} (+7.97)               &    \textbf{60.35} (+9.64)                    \\ \hline
\multirow{4}{*}{CIFAR10}  & \multirow{2}{*}{PINN (baseline)} & \multirow{2}{*}{88.03} & I-FGSM & 46.02                  & 37.03                   & 33.02                   & 28.84                   \\
                          &                                  &                        & PGD    &  42.72                 &     35.02              &     31.14             &       27.31            \\
                          & \multirow{2}{*}{SP-PINN }  & \multirow{2}{*}{87.46} & I-FGSM & \textbf{51.76} (+5.74) & \textbf{51.66} (+14.63) & \textbf{51.56} (+18.54) & \textbf{51.13} (+22.29) \\
                          &                                  &                        & PGD    & \textbf{50.93} (+8.21) & \textbf{50.62} (+15.60) & \textbf{49.94} (+18.80) & \textbf{49.11} (+21.80) \\ \hline
\multirow{4}{*}{CIFAR100} & \multirow{2}{*}{PINN (baseline)} & \multirow{2}{*}{64.47}      & I-FGSM &  23.67                      &         17.03                &         13.90                &    11.70                     \\
                          &                                  &                        & PGD    &  21.00                      &      15.51                   &      12.96                   &     10.89                    \\
                          & \multirow{2}{*}{SP-PINN }  & \multirow{2}{*}{62.32}      & I-FGSM &   \textbf{26.41} (+2.74)                    &      \textbf{26.39} (+9.36)                   &      \textbf{26.27} (+12.37)                &    \textbf{26.16} (+14.46)                  \\
                          &                                  &                        & PGD    &   \textbf{25.93} (+5.93)                   &      \textbf{25.85} (+10.34)                  &     \textbf{25.80} (+12.84)                  &    \textbf{25.48} (+14.59)                 \\ \hline
\end{tabular}
\caption{Classification accuracy (\%) on MNIST, SVHN, and CIFAR. Results that surpass the baseline model are bold. The performance gain in parentheses is compared with the baseline model.}
\label{exp:main}
\end{table*}

\begin{figure*}[t] 
\centering
\begin{center}
\subfigure{\includegraphics[scale=0.28]{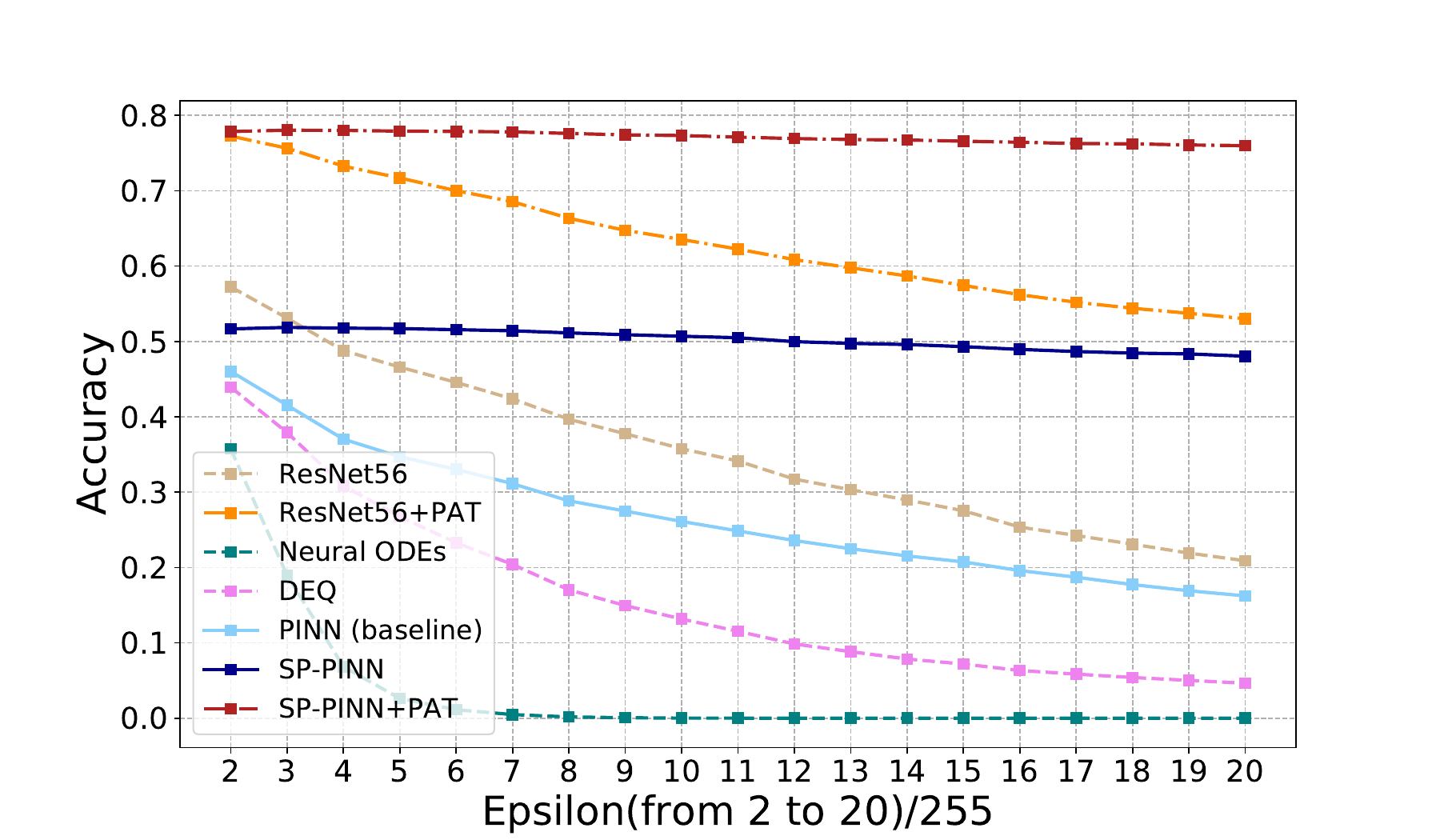}
  \label{1}}
\quad
\subfigure{\includegraphics[scale=0.28]{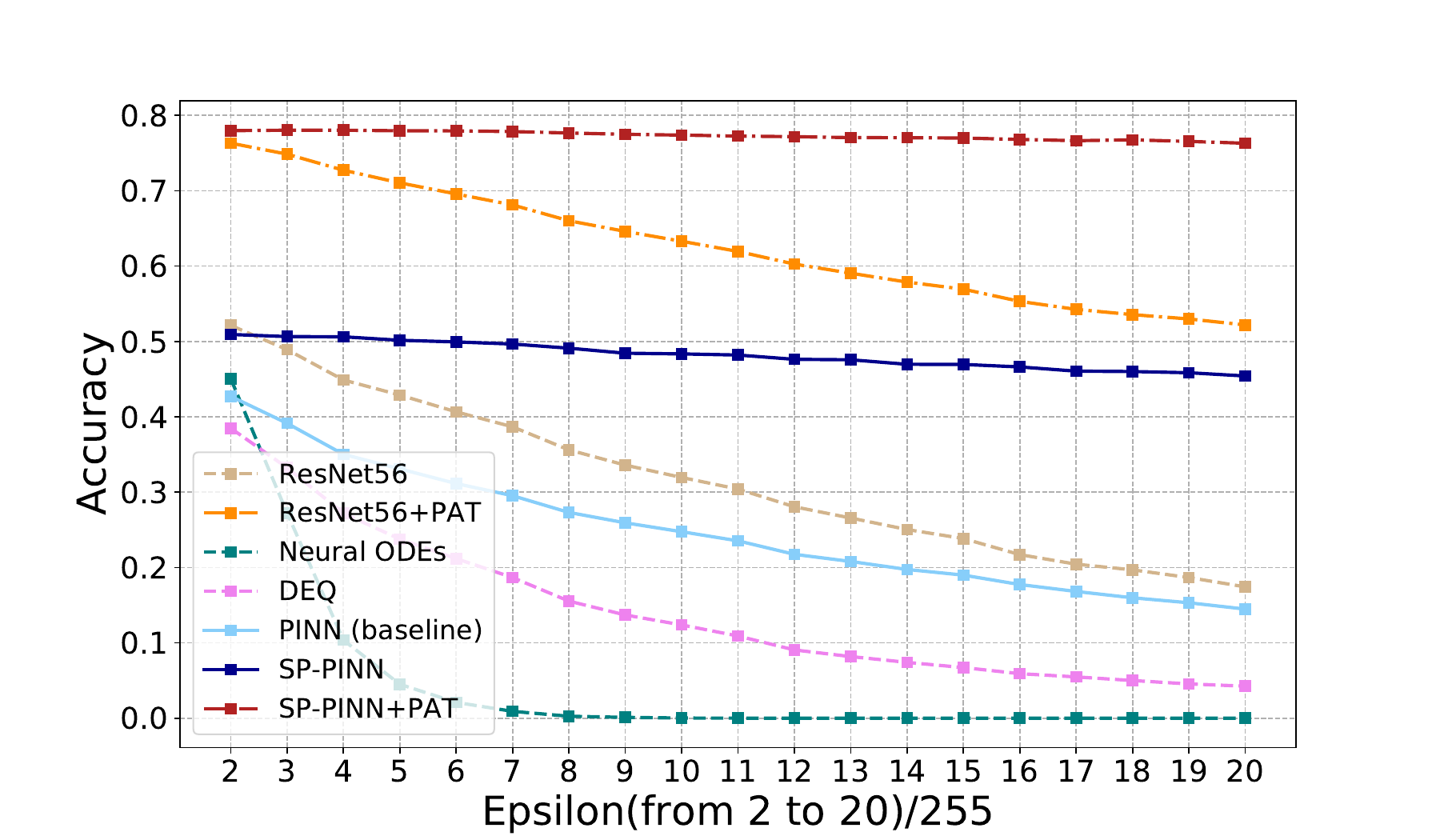}
 \label{2}}
\end{center}
\caption{Comparison between different methods on CIFAR10 under I-FGSM attacks (left) and PGD attacks (right).}
\label{fig:compa}
\end{figure*}

\subsubsection{The Evaluation of the SP-PINN on Image Data}

Table \ref{exp:main} presents the experimental results in terms of classification accuracy and robustness against adversarial examples. On clean data, SP-PINN performs slightly inferior to the baseline model. Figure \ref{exp:time} illustrates that the PINN and SP-PINN are significantly less time-consuming than Neural ODEs. In terms of robustness against adversarial examples, we evaluate the effectiveness of SP-PINN against white-box attacks with varying attack radii from $\epsilon= 2/255$ to $\epsilon= 8/255$. The experimental results demonstrate that the SP-PINN outperforms the baseline model on all datasets. 


Furthermore, we observe that the accuracy improvement under adversarial attacks increases as the attack radii increase for all datasets. For example, the SP-PINN under I-FGSM with attack radii $\epsilon= 2/255$, $\epsilon= 4/255$, $\epsilon= 6/255$ and $\epsilon= 8/255$ achieves a boost of 8.21\%, 15.60\%, 18.80\%, 21.80\% on CIFAR10, respectively. These findings confirm the significant enhancement of robustness achieved by the structure-preserving module in PINN.


\begin{table}[t]
\centering
\begin{tabular}{lllll}
\hline
Radius                            & Attack & +PAT & +RD   & +IAT           \\ \hline
\multirow{2}{*}{$\epsilon=2/255$} & I-FGSM & \underline{77.85}   & 61.61 & \textbf{80.97} \\
                                  & PGD    & \underline{78.04}   & 61.58 & \textbf{80.11} \\ \hline
$\epsilon=4/255$                  & I-FGSM & \underline{78.00}   & 61.57 & \textbf{80.80} \\
                                  & PGD    & \underline{77.98}   & 61.45 & \textbf{79.96} \\ \hline
$\epsilon=6/255$                  & I-FGSM & \underline{77.86}   & 61.54 & \textbf{80.63} \\
                                  & PGD    & \underline{77.95}   & 61.42 & \textbf{79.93} \\ \hline
$\epsilon=8/255$                  & I-FGSM & \underline{77.60}   & 61.46 & \textbf{80.52} \\
                                  & PGD    & \underline{77.67}   & 61.35 & \textbf{79.85} \\ \hline
\end{tabular}
\centering
\caption{Classification accuracy of the SP-PINN combined with adversarial training method on CIFAR10 under adversarial attacks. The second best result is with the underline.}
\label{exp:at}
\end{table}

\subsubsection{SP-PINN With Adversarial Training}

Our approach is independent of other adversarial defense methods, such as adversarial training. This means that we can combine SP-PINN with adversarial training techniques to further enhance defense performance. We consider three well-known adversarial training methods, namely PAT~\cite{madry2017towards}, robust dataset (RD)~\cite{ilyas2019adversarial}, and interpolated adversarial training (IAT)~\cite{LAMB2022218}. 

From Table \ref{exp:at}, we observe that training SP-PINN with adversarial training methods further improves its robustness against adversarial examples. For instance, when training  SP-PINN with PAT, RD, and IAT shows a 26.09\%, 9.85\%, and 29.21\% boost respectively on CIFAR10 under I-FGSM attack with $\epsilon= 2/255$. 

Figure \ref{fig:compa} provides a comparison between ResNet56 \cite{he2016deep}, ResNet56 with PAT, deep equilibrium model (DEQ) \cite{bai2019deep}, Neural ODEs \cite{chen2018neural}, PINN, SP-PINN, and SP-PINN with PAT under adversarial attacks ranging from $\epsilon=2/255$ to $\epsilon=20/255$. It is apparent that SP-PINN is insensitive to the radius of adversarial attack, which further corroborates the effectiveness of our method.

\section{Conclusions}

In this paper, we proposed SP-PINNs by introducing the prior knowledge about the properties of the underlying dynamical systems. The applicability of the proposed SP-PINNs ranges from PDEs, and to image recognition. Experimental results showed that our approach improved the accuracy of the numerical solutions in solving the Allen–Cahn equation. Specially, on image recognition tasks, SP-PINN demonstrated high robustness against adversarial examples. Our findings highlighted the power of PINN for handling image data. In future work, we will consider employing the proposed method for solving more complex `conservative' and `dissipative' equations. Additionally, we will refine our model and enhance its capability to handle large-scale datasets.

\section*{Ethical Statement}
There are no ethical issues.
\bibliographystyle{named}
\bibliography{ijcai24}

\end{document}